\documentclass[a4paper, 10pt, conference]{ieeeconf}
\usepackage{FG2027}
\FGfinalcopy
\IEEEoverridecommandlockouts
\usepackage[utf8]{inputenc}
\usepackage[T1]{fontenc}
\usepackage{graphicx}
\usepackage{booktabs}
\usepackage{multirow}
\usepackage{amsmath}
\usepackage{xcolor}
\usepackage{url}
\graphicspath{{figures_matched/}{./}{figures/}}

\begin{document}

\title{\LARGE \bf Pulse or Provenance? What rPPG-Based Deepfake Detectors Learn}
\author{Othmane Harraq\\ Temple University\\ {\tt\small othmane.harraq@temple.edu} \and Tamer Aldwairi\\ Temple University\\ {\tt\small aldwairi@temple.edu}}
\maketitle
\thispagestyle{plain}
\pagestyle{plain}

\begin{abstract}
Remote photoplethysmography (rPPG) has been proposed as a physiological cue for deepfake detection, based on the premise that synthesized faces lack a genuine cardiac signal. We examine what rPPG-style detectors learn on Celeb-DF++ (22 forgery methods) and FaceForensics++ (FF++), using one identically sampled window per real and fake video. On talking-face forgeries, simple chrominance traces reach an AUC of 0.998, exceeding a dedicated rPPG extractor (0.87). Controlled experiments indicate that this signal is not physiological in origin. On fully synthesized forgeries, it is equally discriminative when extracted from background regions or from full frames without face detection, and it is concentrated above the 0.7 to 4\,Hz heart-rate band. It is largely removed by shuffling frame order but persists after normalizing resolution, frame rate and encoding, and an independent reimplementation reproduces these findings. Removing high-frequency content makes most synthesized videos appear real, adding it makes real videos appear fake, and injecting a synthetic pulse has little effect. In controlled self-reenactment experiments with two generators, animating a face from a single frame introduces the same high-frequency signal, whereas reconstructing each frame from its own source does not. Face-based chrominance detectors trained on Celeb-DF++ classify over 99\% of real FF++ videos as fake, indicating that they largely encode video source rather than manipulation, and a CLIP-based detector exhibits the same behavior. A frozen CLIP backbone matches or exceeds fusions of CLIP with rPPG and frequency features. Finally, we show that the official Celeb-DF and Celeb-DF++ splits share identities and source-video content between training and test, and that custom protocols reusing test videos and identities in training inflate weaker detectors by up to 0.30 AUC.
\end{abstract}

\section{Introduction}

\begin{figure}[t]
\centering
\includegraphics[width=\columnwidth]{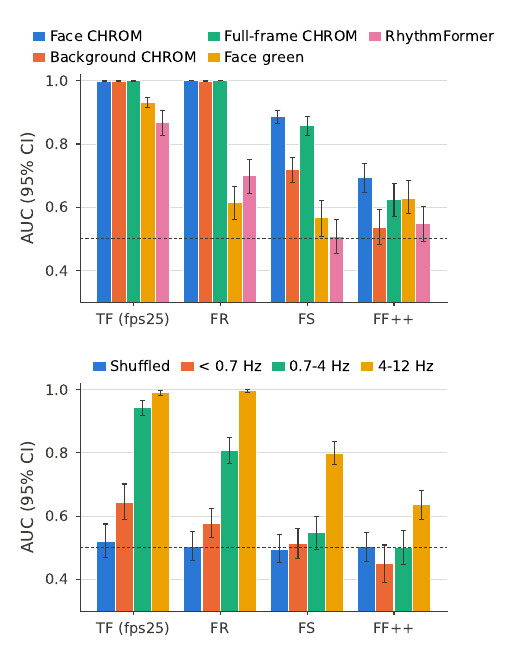}
\caption{rPPG-style detectors do not need a face or a heartbeat. Top: real vs.\ fake AUC by region and signal. Bottom: face CHROM with frames shuffled or restricted to one frequency band. On talking-face (TF) and face-reenactment (FR) forgeries, the background and the full frame separate real from fake as well as the face, the 4--12\,Hz band beats the 0.7--4\,Hz heart-rate band, and shuffling frames removes the signal. Matched sampling, test identities; dashed line is chance.}
\label{fig:teaser}
\end{figure}

Physiological signals are an attractive basis for deepfake detection because they come with a mechanistic argument. Remote photoplethysmography (rPPG) recovers the small, periodic color changes that blood flow produces in facial skin \cite{verkruysse2008,dehaan2013,wang2017pos}, and a synthesized face has no blood flow. Detectors built on this premise, including FakeCatcher \cite{ciftci2020}, DeepRhythm \cite{qi2020}, and DeepFakesON-Phys \cite{hernandez2021}, report high accuracy on face-swap benchmarks. The same reasoning underlies pulse-based face anti-spoofing \cite{li2016pulse,liu2018fas}.

The premise is strongest for talking-face (TF) forgeries, which animate a single still image from audio and contain no real video from which a pulse could be inherited. Face-swap forgeries, by contrast, can carry the pulse of their source footage \cite{seibold2025}. TF forgeries are therefore the setting in which an rPPG detector should work best and for the reasons its designers intend.

On the TF subset of Celeb-DF++ \cite{li2025celebdfpp}, the result is puzzling. A plain chrominance (CHROM) trace \cite{dehaan2013} separates real from TF videos with an AUC of 0.998 (Fig.~\ref{fig:teaser}), while a state-of-the-art rPPG extractor, RhythmFormer \cite{zou2025rhythmformer}, reaches only 0.87. If the discriminative information were a heartbeat, a cruder pulse estimate should not outperform a better one.

Prior work asked whether forgeries carry a valid pulse \cite{damelio2023,seibold2025}; we ask what the detectors use instead and where it comes from. Under a matched sampling protocol, in which every real and fake video contributes one identically positioned window, we test where the signal lives, what kind of signal it is, whether it survives format normalization, how it responds to targeted interventions, and what it says about the video it is computed from. Our findings are:

\begin{itemize}
\item \textbf{The signal is not primarily physiological.} On fully synthesized forgeries it is as strong in background regions and in full frames without face detection as on the face. It is concentrated above the 0.7 to 4\,Hz heart-rate band, disappears when frame order is shuffled, and survives normalization of resolution, frame rate, and encoding. An independent reimplementation reproduces these results.
\item \textbf{It is introduced by image-to-video synthesis.} In controlled self-reenactment experiments with two generators, animating a face from a single frame of a real video adds the same high-frequency signal, while reconstructing each frame from itself does not.
\item \textbf{The detectors largely encode video source.} Face-based chrominance detectors trained on Celeb-DF++ label over 99\% of real FF++ videos as fake, and the same features separate two collections of real videos. A CLIP-based detector shows the same behavior.
\item \textbf{Evaluation practice obscures this.} The official Celeb-DF and Celeb-DF++ splits share identities and source content between training and test, and custom protocols that reuse test videos inflate weak detectors by up to 0.30 AUC.
\end{itemize}

We close with a short set of controls that physiological detectors should report. Code and evaluation splits will be made publicly available.

\section{Related Work}

\textbf{Physiological signals for liveness and deepfake detection.} rPPG estimation ranges from classical color-space projections such as CHROM \cite{dehaan2013} and POS \cite{wang2017pos} to learned extractors such as RhythmFormer \cite{zou2025rhythmformer}. Pulse cues were first used for face anti-spoofing, where a printed photo or a replayed screen should show no pulse \cite{li2016pulse,liu2018fas,hernandez2018pad}, and later for deepfake detection \cite{ciftci2020,qi2020,hernandez2021,liang2021}. Harraq and Aldwairi \cite{harraq2026pulseguard} applied rPPG-derived signals to TF forgeries but did not establish that the discriminative content was cardiac, and later combined these signals with lip-region DCT features \cite{harraq2026dct}. Two studies question the premise directly. D'Amelio et al.\ \cite{damelio2023} found that rPPG-based detection on face-swap data does not track signal quality as expected, and Seibold et al.\ \cite{seibold2025} showed that face-swap forgeries can inherit a valid pulse from their driver video. We study the opposite case, forgeries with no driver video, and identify what the detectors use instead.

\textbf{Temporal artifacts in video forgeries.} A separate line of work detects forgeries from temporal inconsistency without appealing to physiology, for example through mouth motion \cite{haliassos2021lips}, short-range temporal coherence \cite{zheng2021ftcn}, or inconsistencies across multiple time scales \cite{yu2025multiscale,guo2025temporal}. Our findings place rPPG-style detectors in this group: the signal they use is a temporal color artifact, measured through a physiological lens.

\textbf{Shortcuts and dataset bias.} Classifiers often exploit features that are predictive in a dataset but unrelated to the intended task \cite{geirhos2020}. In deepfake detection, a leading silence in fake audio was shown to separate real and fake samples in two widely used audio-visual datasets \cite{smeu2025silence}, compression-format differences bias generated-image detectors \cite{grommelt2024}, and binary detectors absorb identity information \cite{dong2023iid}. Our contribution is to show that a detector class justified by biology is largely a source and synthesis detector, and to measure this with controls specific to temporal signals.

\textbf{Benchmarks and splits.} Celeb-DF \cite{li2020celebdf} and its extension Celeb-DF++ \cite{li2025celebdfpp} provide 59 celebrities and, in Celeb-DF++, 22 forgery methods across face-swap (FS), face-reenactment (FR), and TF categories, with an official test list. FF++ \cite{rossler2019} defines splits by pairs of source videos. We examine how these splits separate identities and content.

\section{Experimental Setup}

\subsection{Data and splits}
We use all 22 methods of Celeb-DF++ \cite{li2025celebdfpp} and the five manipulation methods of FF++ (c23) \cite{rossler2019}. For Celeb-DF++, we partition the 59 identities into 41 training, 9 validation, and 9 test identities. FS and FR filenames name two identities; a fake is assigned to a split only if both identities belong to it, and is excluded otherwise. TF filenames name one celebrity. For FF++ we use the official split. Section~\ref{sec:eval} additionally analyzes the official Celeb-DF and Celeb-DF++ test lists. All reported results use test identities only.

\subsection{Matched temporal sampling}
Real and fake videos differ in length and frame rate. To prevent the sampling procedure from becoming a cue, every video, real or fake, contributes exactly one window: its first 160 frames. Videos shorter than 160 native frames are excluded rather than resampled; this removes all HyperReenact videos, so that method has no matched result. The length is the training chunk length of the RhythmFormer UBFC configuration and was inherited, not tuned. On TF, face CHROM reaches 0.998 at 80, 128, and 160 frames, the band, shuffle, and background results hold at each length, and selection on validation identities also picks 160 (Section~\ref{sec:notphys}). Real Celeb-DF++ videos are 30\,fps, while most generators output 25\,fps; frequency analyses therefore use the fps25 variant, in which real videos are downsampled to 25\,fps by frame dropping and both classes are re-encoded identically.

\subsection{Signals and regions}
For each window we compute per-frame mean RGB values over three regions. The face region is one MediaPipe BlazeFace \cite{bazarevsky2019} box per video, enlarged 1.5$\times$ and fixed for all frames. The background region is a square of the same size at the frame corner farthest from the face, if it does not overlap the enlarged box; otherwise, as in fake videos whose face fills the frame, it is the mean over all pixels outside the box. The full-frame region is the whole image, with no face detection. From each RGB series we derive three traces. The green trace is the green channel. CHROM \cite{dehaan2013} normalizes each channel by its temporal mean, forms $X = 3R - 2G$ and $Y = 1.5R + G - 1.5B$, and returns $X - \alpha Y$ with $\alpha = \sigma(X)/\sigma(Y)$. POS \cite{wang2017pos} projects the normalized channels onto the plane orthogonal to skin tone and combines the two projections with a standard-deviation ratio. We also run RhythmFormer \cite{zou2025rhythmformer} (UBFC checkpoint) on the face crops. Each trace is z-scored per window before classification.

\textbf{Frequency bands.} To isolate frequency ranges, we zero the real FFT coefficients of the raw window outside a band, at the video's frame rate, and then z-score the result. The bands are below 0.7\,Hz, 0.7--4\,Hz (the usual heart-rate range), and 4--12\,Hz, capped below the 12.5\,Hz Nyquist frequency of 25\,fps video so that both classes are filtered identically.

\subsection{Classifiers}
Traces are classified with a small 1D ResNet with three residual stages and dropout 0.5, trained for a fixed 30 epochs with AdamW (learning rate $10^{-3}$, weight decay $5\times10^{-4}$, batch 64) and a class-weighted binary cross-entropy loss. As general-purpose references we use small MLP heads on frozen CLIP ViT-L/14 \cite{radford2021clip} and Xception \cite{chollet2017} features, averaged over 10 frames of the same window (frames 0, 16, \dots, 144), and fusions of CLIP with rPPG and lip-region DCT features \cite{harraq2026dct}. The rPPG+DCT detector concatenates MLP embeddings of the face RhythmFormer waveform and a 64-coefficient DCT descriptor of a 64$\times$64 lip patch (23K parameters); the fusion projects CLIP, rPPG, and DCT features to 256 dimensions each and combines them with learned branch weights. Validation identities are used only for early stopping of these heads; the trace classifiers do not use them.

\subsection{Evaluation}
We report video-level AUC as the mean over 5 seeds with 95\% bootstrap confidence intervals (1,000 resamples of test videos), and identity-level bootstrap intervals for headline results. Because real test videos are outnumbered by fakes (up to 1:6), we also report precision-recall metrics: for TF face CHROM, PR-AUC is 0.998 with fakes as the positive class and 0.999 with reals as the positive class, the EER is 1.0\%, and AUC on class-balanced subsamples is 0.998. For TF RhythmFormer they are 0.770, 0.914, 19.7\%, and 0.869, so weaker detectors look worse under these metrics than under AUC. Face detection fails for at most 1\% of Celeb-DF++ videos and 2--3\% of FF++ videos in either class; the 160-frame rule excludes 43\% of TF fakes and 9\% of FS and FR fakes but almost no real videos. Random-label controls are reported as the distribution over 20 label draws. All runs are deterministic. An independent reimplementation that shares no code with the main pipeline reproduces the main results (Section~\ref{sec:notphys}).

\section{The Signal Is Not Primarily Physiological}
\label{sec:notphys}

\begin{table*}[t]
\centering
\caption{Real vs.\ fake AUC under matched sampling (one window per video, test identities, mean of 5 seeds). Shuffled and band columns use the face signal. Bands are applied before per-window normalization. TF uses the fps25 variant. Counts are fake/real test videos. RhythmFormer operates only on detected face crops, so it has no background or full-frame entries (--).}
\label{tab:controls}
\begin{tabular}{llcccccc}
\toprule
Set & Signal & Face & Background & Full frame & Shuffled & 0.7--4\,Hz & 4--12\,Hz \\
\midrule
\multirow{4}{*}{TF (384/98)} & CHROM & 0.998 & 0.998 & 0.999 & 0.521 & 0.943 & 0.992 \\
            & POS   & 0.998 & 0.993 & 0.998 & 0.593 & 0.897 & 0.970 \\
            & Green & 0.931 & 0.960 & 0.959 & 0.604 & 0.802 & 0.911 \\
            & RhythmFormer & 0.869 & -- & -- & 0.679 & 0.860 & 0.800 \\
\midrule
\multirow{4}{*}{FR (315/99)} & CHROM & 1.000 & 0.998 & 1.000 & 0.505 & 0.809 & 0.996 \\
            & POS   & 1.000 & 0.995 & 1.000 & 0.493 & 0.740 & 0.994 \\
            & Green & 0.614 & 0.915 & 0.853 & 0.551 & 0.602 & 0.672 \\
            & RhythmFormer & 0.701 & -- & -- & 0.484 & 0.662 & 0.695 \\
\midrule
\multirow{4}{*}{FS (568/99)} & CHROM & 0.888 & 0.720 & 0.860 & 0.496 & 0.547 & 0.798 \\
            & POS   & 0.877 & 0.750 & 0.895 & 0.482 & 0.507 & 0.790 \\
            & Green & 0.568 & 0.660 & 0.609 & 0.515 & 0.519 & 0.518 \\
            & RhythmFormer & 0.509 & -- & -- & 0.465 & 0.523 & 0.503 \\
\midrule
\multirow{4}{*}{FF++ (207/139)} & CHROM & 0.694 & 0.537 & 0.623 & 0.504 & 0.500 & 0.637 \\
            & POS   & 0.684 & 0.555 & 0.620 & 0.527 & 0.558 & 0.645 \\
            & Green & 0.629 & 0.492 & 0.535 & 0.512 & 0.540 & 0.587 \\
            & RhythmFormer & 0.549 & -- & -- & 0.506 & 0.567 & 0.573 \\
\bottomrule
\end{tabular}
\end{table*}

\subsection{Near-perfect detection from a crude signal}
On TF forgeries, a CHROM trace from the face separates real and fake videos with an AUC of 0.998 [0.997, 1.000] (Table~\ref{tab:controls}). POS performs identically. RhythmFormer, a dedicated rPPG extractor, reaches 0.869 [0.827, 0.907], and a green-channel trace 0.931. The ordering is the reverse of what a physiological explanation predicts: the signals that are worst at estimating a pulse are best at detection. The identity-level bootstrap interval for CHROM, [0.996, 1.000], is as narrow as the video-level one.

\subsection{The face is not needed}
On fully synthesized forgeries the signal does not depend on skin. A background region gives 0.998 on TF and 0.998 on FR, and the mean color of the full frame, computed without any face detection, gives 0.999 and 1.000 (Table~\ref{tab:controls}). Even the green channel is stronger from the background than from the face (0.960 vs.\ 0.931 on TF, 0.915 vs.\ 0.614 on FR). Face-detector instability is not the cause: our traces use one box per video, and box jitter measured per frame is lower in fake videos than in real ones. On FS and FF++, which keep real footage outside the face, the background is much weaker than the face (0.720 vs.\ 0.888 and 0.537 vs.\ 0.694). We return to this difference in Section~\ref{sec:origin}.

\subsection{The signal is temporal and lies above the heart-rate band}
Shuffling the frames within each window removes the signal: CHROM falls to 0.521 on TF and 0.505 on FR, and to chance on FS and FF++. Static color statistics of the same windows carry no information (single-feature AUC 0.48--0.55). Among simple per-window statistics, only temporal ones separate the classes: fake traces vary less over the window as a whole (temporal standard deviation, AUC 0.25) but are less smooth from frame to frame (AUC 0.905 for a smoothness score). The discriminative content is therefore in how color changes from frame to frame.

\begin{figure*}[t]
\centering
\includegraphics[width=0.96\textwidth]{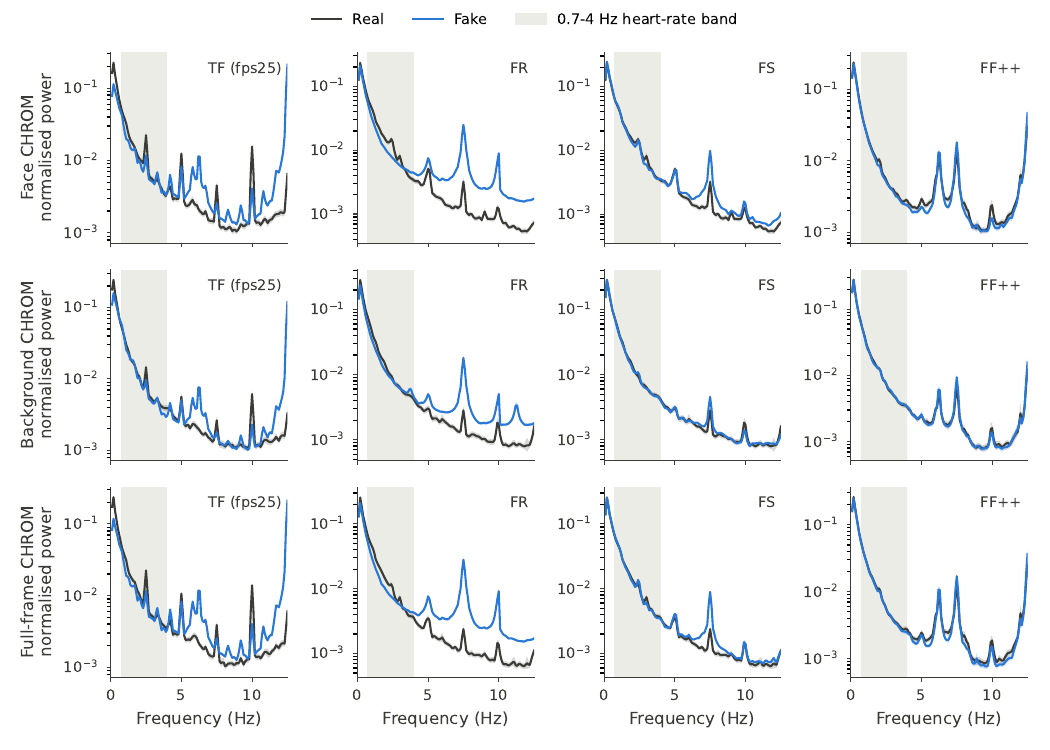}
\caption{Averaged power spectra of real (black) and fake (blue) CHROM traces under matched sampling. Rows: face, background, full frame. Columns: forgery category. The shaded band is 0.7--4\,Hz. Real and fake spectra separate above 4\,Hz for TF and FR, only slightly for FS, and not at all for FF++.}
\label{fig:spectra}
\end{figure*}

To locate it in frequency, we filter each window into three bands before normalization. On TF, the 4--12\,Hz band alone reaches 0.992, the 0.7--4\,Hz heart-rate band 0.943, and the band below 0.7\,Hz 0.644. The high band exceeds the heart-rate band in every set we tested, for CHROM and POS, and at every window length (80, 128, 160 frames). The power spectra agree (Fig.~\ref{fig:spectra}): on TF, 17.1\% [16.8, 17.5] of the face CHROM trace power lies in 4--12\,Hz for fakes against 12.1\% [11.4, 12.7] for reals, and on FR the gap is larger (19.4\% vs.\ 6.3\%). On FS the excess is small (8.4\% vs.\ 6.3\%), and on FF++ it is absent (13.1\% vs.\ 13.4\%). RhythmFormer is the exception, as expected for a model trained to output pulse-band waveforms.

\subsection{The signal is not a file-format artifact}

\begin{table}[t]
\centering
\caption{TF face CHROM under progressively stronger normalization (matched sampling), and an independent reimplementation (logistic regression on the trace spectrum, no shared code). Band columns are N/A for native files, whose frame rates differ between classes (25 vs.\ 30\,fps), and for shuffled traces, which have no temporal structure to filter.}
\label{tab:ladder}
\begin{tabular}{lccc}
\toprule
Variant & Plain & 0.7--4\,Hz & 4--12\,Hz \\
\midrule
Native files & 0.998 & N/A & N/A \\
fps25 (frame rate, codec matched) & 0.998 & 0.943 & 0.992 \\
res\_norm (face crop 256$^2$, 30\,fps) & 0.998 & 0.859 & 0.993 \\
res25 (crop, resolution, fps, codec) & 0.986 & 0.906 & 0.978 \\
\midrule
Reimplementation, face & 0.989 & 0.802 & 0.983 \\
Reimplementation, background & 0.995 & 0.698 & 0.952 \\
Reimplementation, shuffled & 0.531 & N/A & N/A \\
\bottomrule
\end{tabular}
\end{table}

In the native files, real and TF videos differ in codec, frame rate, keyframe interval, resolution, and audio. Re-encoding both classes to the same frame rate and codec, cropping and resizing faces to a common resolution, and combining all of these leaves the AUC at 0.986 or higher (Table~\ref{tab:ladder}); after full normalization no metadata field separates the classes. Notching out the frequencies of the real videos' original keyframe interval changes the AUC no more than an equally wide control notch (0.995 vs.\ 0.999). Heavier compression weakens but does not remove the cue: at CRF 40, models trained on standard encodes reach 0.857 [0.833, 0.882], and 0.931 [0.913, 0.947] after retraining. A standalone reimplementation, using a different classifier and no shared code, reproduces every conclusion.

\subsection{Interventions}

\begin{table}[t]
\centering
\caption{Interventions on test traces with unmodified TF face-CHROM models (fps25, matched sampling). Flipped: fraction of the modified class on the wrong side of the unmodified models' EER threshold. Low-pass filters also remove content up to the 12.5\,Hz Nyquist limit.}
\label{tab:interv}
\begin{tabular}{lcc}
\toprule
Intervention & AUC & Flipped \\
\midrule
None & 0.998 & 2.0\% of each class \\
Fakes low-passed at 8\,Hz & 0.589 & 97\% of fakes \\
Fakes low-passed at 4\,Hz & 0.877 & 85\% of fakes \\
Fakes, 12--12.5\,Hz removed & 0.825 & 75\% of fakes \\
Fakes, 4--12\,Hz removed & 0.999 & 1.6\% of fakes \\
Reals + 4--12\,Hz noise (fake level) & 0.998 & 1\% of reals \\
Reals + 12--12.5\,Hz noise (fake level) & 0.510 & 99.8\% of reals \\
Fakes + synthetic pulse & 0.996 & 2.5\% of fakes \\
Reals, 0.7--4\,Hz removed & 0.996 & 3.7\% of reals \\
Reals, 4.5--7.8\,Hz removed (control) & 0.990 & 9.0\% of reals \\
Reals, 7.8--11.1\,Hz removed (control) & 0.990 & 15.7\% of reals \\
\bottomrule
\end{tabular}
\end{table}

Removing high-frequency content from fake traces makes nearly all of them appear real: after low-passing at 8\,Hz, 97\% of fakes are called real, against 2.0\% without modification (Table~\ref{tab:interv}). The models rely most on content nearest the 12.5\,Hz Nyquist limit, that is, on alternation between consecutive frames. Removing only 12--12.5\,Hz makes 75\% of fakes appear real, while removing 4--12\,Hz leaves them detected (1.6\%), because the near-Nyquist content that remains is enough. Low-passing at 4\,Hz flips fewer fakes than at 8\,Hz because it also makes them smoother than real traces (mean absolute frame-to-frame change 0.253 vs.\ 0.426), which the models treat as atypical. This content is also sufficient. TF fakes carry 26\% of their trace power in 12--12.5\,Hz against 1.4\% for reals; raising real traces to the fake level in this band makes 99.8\% of them appear fake (AUC 0.510), whereas fake-level noise spread over 4--12\,Hz does not (1\%). Injecting a synthetic pulse into fake traces, a 1.0--1.7\,Hz sinusoid with the mean heart-rate-band power of real traces, leaves 2.5\% [1.7, 3.4] of them called real, close to the 2.0\% baseline, and doubling its power gives 6.2\%. The presence of a cardiac component is therefore not what makes a video look real to these models. Removing 0.7--4\,Hz from fake traces changes nothing (2.2\% called real). Conversely, removing the 0.7--4\,Hz band from real traces flips only 3.7\% [1.6, 6.3] of reals, fewer than removing an equally wide band outside the heart-rate range (9.0\% for 4.5--7.8\,Hz, 15.7\% for 7.8--11.1\,Hz). The chrominance detectors are therefore no more sensitive to the heart-rate band than to arbitrary bands of the same width. RhythmFormer is the exception: removing the heart-rate band flips 13.7\% of reals against 5.1\% and 1.4\% for the control bands, consistent with an extractor trained to output pulse-band waveforms. It is also the weakest of the four detectors (Table~\ref{tab:controls}).

\section{Where the Signal Comes From}
\label{sec:origin}

\subsection{Controlled self-reenactment}
The previous section shows what the signal is not. To test whether synthesis introduces it, we regenerate 199 real Celeb-DF videos from themselves with two Celeb-DF++ reenactment methods, LivePortrait \cite{guo2024liveportrait} and the Thin-Plate Spline Motion Model (TPSMM) \cite{zhao2022tpsmm}. Each reenactment shares its person, scene, motion, and final encoding with its source video. In mode (a) the face is animated from the first frame and pasted into it; in mode (b) the full face crop is generated; in mode (a2) each frame is reconstructed from itself.

\begin{figure}[t]
\centering
\includegraphics[width=\columnwidth]{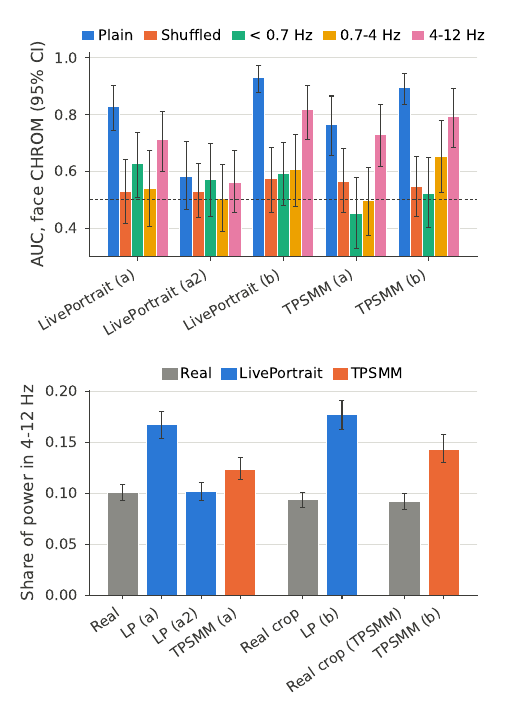}
\caption{Real videos vs.\ their own self-reenactments. Top: face CHROM AUC for separating each reenactment from its source video, plain, shuffled, and per frequency band. Bottom: share of face CHROM power in 4--12\,Hz for each real video and its reenactments. (a) animated from the first frame and pasted back; (b) generated face crop; (a2) each frame rebuilt from itself. LP: LivePortrait.}
\label{fig:mechanism}
\end{figure}

Animating a face from a single frame makes the reenactment separable from its own source video, under the same matched sampling as before (Fig.~\ref{fig:mechanism}). For LivePortrait, face CHROM reaches 0.830 [0.744, 0.902] in mode (a) and 0.931 [0.878, 0.971] in mode (b); for TPSMM, 0.767 [0.656, 0.865] and 0.896 [0.835, 0.944]. Both show the signature of the Celeb-DF++ forgeries: the 4--12\,Hz band exceeds the heart-rate band (LivePortrait 0.711 vs.\ 0.539 and 0.818 vs.\ 0.607; TPSMM 0.728 vs.\ 0.497 and 0.793 vs.\ 0.652), the share of high-band power rises above that of the source video (LivePortrait 16.7\% and 17.8\% vs.\ about 10\%; TPSMM 12.4\% and 14.3\%), and shuffling removes the effect. Reenactments also carry more power just below the Nyquist limit than their sources (at 30\,fps, 14.5--15\,Hz: LivePortrait 10.0\% and 11.3\% vs.\ about 3.5\%; TPSMM 5.7\% and 6.3\%), though far less than TF fakes do. The effect is weaker for TPSMM than for LivePortrait. Reconstructing each frame from itself with LivePortrait adds almost nothing (0.582 [0.465, 0.707]; high-band power 10.2\% vs.\ 10.1\%); TPSMM has no frame-by-frame mode, so this control uses LivePortrait only. The cue is thus introduced by synthesizing a sequence from a still appearance, not by the rendering and pasting pipeline. In mode (a), the frozen background is also separable, which reflects the absence of motion rather than added content, and we do not count it as evidence. The test split contains 32 to 34 clips per class, so intervals are wide.

\begin{table}[t]
\centering
\footnotesize
\setlength{\tabcolsep}{4pt}
\caption{Per-method real vs.\ fake AUC under matched sampling (mean of 5 seeds; test identities; about 98--139 real test videos per set). Face, Bg and Frame are CHROM traces from the face, background and full frame; RF is RhythmFormer. $^\dagger$EchoMimic outputs 24\,fps and is scored on native files. Celeb-DF denotes the original Celeb-DF-v2 forgeries.}
\label{tab:permethod}
\begin{tabular}{llccccc}
\toprule
 & Method & $n_\text{fake}$ & Face & Bg & Frame & RF \\
\midrule
FS & BlendFace & 55 & 0.945 & 0.772 & 0.911 & 0.534 \\
 & Celeb-DF & 183 & 0.604 & 0.509 & 0.629 & 0.473 \\
 & GHOST & 55 & 0.973 & 0.756 & 0.923 & 0.502 \\
 & HifiFace & 55 & 0.953 & 0.769 & 0.907 & 0.539 \\
 & InSwapper & 55 & 0.954 & 0.749 & 0.905 & 0.482 \\
 & MobileFaceSwap & 55 & 0.721 & 0.507 & 0.695 & 0.490 \\
 & SimSwap & 55 & 1.000 & 0.950 & 0.998 & 0.528 \\
 & UniFace & 55 & 0.951 & 0.747 & 0.909 & 0.524 \\
\midrule
FR & DaGAN & 52 & 1.000 & 0.999 & 1.000 & 0.690 \\
 & FSRT & 52 & 1.000 & 0.998 & 1.000 & 0.724 \\
 & HyperReenact & \multicolumn{5}{c}{no video $\geq$160 frames} \\
 & LIA & 52 & 1.000 & 0.999 & 1.000 & 0.724 \\
 & LivePortrait & 55 & 1.000 & 0.997 & 1.000 & 0.672 \\
 & MCNET & 52 & 1.000 & 0.998 & 1.000 & 0.682 \\
 & TPSMM & 52 & 1.000 & 0.999 & 1.000 & 0.713 \\
\midrule
TF & AniTalker & 68 & 1.000 & 0.998 & 1.000 & 0.905 \\
 & EchoMimic$^\dagger$ & 50 & 1.000 & 1.000 & 1.000 & 0.786 \\
 & EDTalk & 68 & 1.000 & 0.998 & 1.000 & 0.898 \\
 & FLOAT & 64 & 1.000 & 1.000 & 1.000 & 0.821 \\
 & IP-LAP & 58 & 0.990 & 0.993 & 0.996 & 0.711 \\
 & Real3D-Portrait & 65 & 1.000 & 1.000 & 1.000 & 0.953 \\
 & SadTalker & 61 & 1.000 & 0.999 & 1.000 & 0.926 \\
\midrule
FF++ & Deepfakes & 42 & 0.607 & 0.524 & 0.574 & 0.662 \\
 & Face2Face & 41 & 0.694 & 0.561 & 0.640 & 0.435 \\
 & FaceSwap & 42 & 0.696 & 0.495 & 0.609 & 0.592 \\
 & FaceShifter & 41 & 0.729 & 0.540 & 0.653 & 0.596 \\
 & NeuralTextures & 41 & 0.746 & 0.564 & 0.642 & 0.459 \\
\bottomrule
\end{tabular}
\end{table}

\subsection{Consistency across forgery types}
This account predicts the pattern across forgery types in Table~\ref{tab:controls}, and it holds method by method (Table~\ref{tab:permethod}). TF and FR methods generate the whole frame from a still source and are detected almost perfectly, including from the background. FS methods replace the face in real footage and are detected less well, with a much weaker background. The two FS methods that largely escape are the original Celeb-DF forgeries (0.604), whose generation pipeline was designed to reduce temporal flicker \cite{li2025celebdfpp}, and MobileFaceSwap (0.721). FF++ manipulations edit only the face of real footage; under matched sampling, detection there is weak (0.694), the background is near chance, and fakes carry no excess high-band power. Sampling matters here: when real videos contribute windows from across their full length while fakes contribute their first window, FF++ detection rises to 0.901, and a size-matched control shows that the increase comes from comparing later real windows with the first fake window, not from the amount of training data. RhythmFormer, which outputs pulse-band waveforms, is near chance on FS and FF++.

\section{The Detectors Encode Video Source}
\label{sec:source}

\begin{table}[t]
\centering
\caption{Real videos from other sources, scored by detectors trained on Celeb-DF++ (matched sampling). Called fake: fraction above each model's own EER threshold. Last column: AUC separating held-out Celeb-real from FF++ real using the fake score.}
\label{tab:source}
\setlength{\tabcolsep}{4pt}
\begin{tabular}{llccc}
\toprule
Training & Model & YouTube & FF++ & Celeb vs FF++ \\
\midrule
TF & Face CHROM & 1\% & 99.1\% & 0.998 \\
TF & Background CHROM & 3\% & 77.8\% & 0.959 \\
TF & CLIP & 38\% & 86.0\% & 0.948 \\
All 22 & Face CHROM & 18\% & 99.7\% & 0.999 \\
All 22 & CLIP & 47\% & 87.8\% & 0.945 \\
\bottomrule
\end{tabular}
\end{table}

\begin{figure}[t]
\centering
\includegraphics[width=\columnwidth]{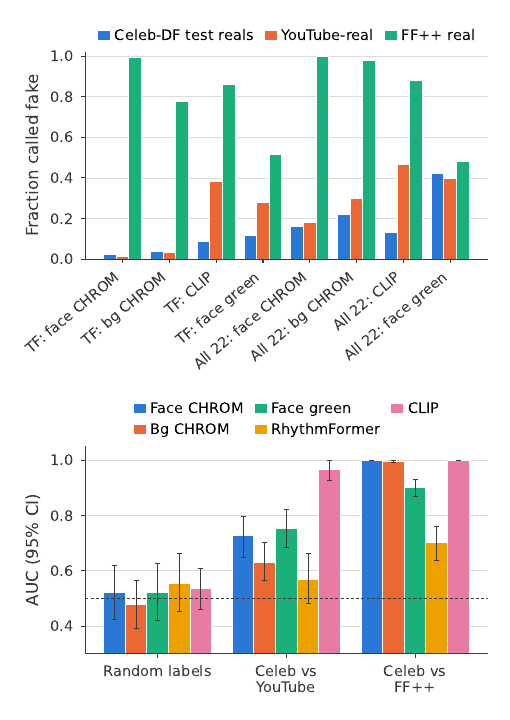}
\caption{The detectors encode video source. Top: fraction of each real collection called fake by detectors trained on the TF subset or on all 22 Celeb-DF++ methods, at each detector's own EER threshold. Bottom: classifiers trained to separate two collections of real videos, with a random-label control.}
\label{fig:source}
\end{figure}

\subsection{Scoring real videos from another source}
If a detector learned a property of forgeries, real videos from a different dataset should be scored as real. Detectors trained on Celeb-DF++ real and fake videos instead label 99\% or more of real FF++ videos as fake (Table~\ref{tab:source}, Fig.~\ref{fig:source}), and their fake score separates held-out Celeb-DF real videos from FF++ real videos almost perfectly. YouTube-real videos, collected through the same pipeline as Celeb-DF real videos, are mostly scored as real by the CHROM detectors. In practical terms, these detectors largely learned that a video not from the Celeb-DF real collection is fake. This is not a generic shift between datasets, since YouTube-real videos, which share the Celeb-DF collection pipeline but not its exact videos, are mostly scored as real. POS-based detectors behave the same way (94--100\% of FF++ reals called fake).

\subsection{Real versus real}
The same features separate two collections of real videos when trained to do so: Celeb-DF real vs.\ FF++ real reaches 1.000 for CHROM and for CLIP, and Celeb-DF real vs.\ YouTube-real reaches 0.727 for CHROM and 0.968 for CLIP. With random labels, the classifiers stay at chance (CHROM 0.515 $\pm$ 0.050, CLIP 0.505 $\pm$ 0.063 over 20 label draws), so this is not a tendency to find differences everywhere.

\subsection{General-purpose detectors}

\begin{table}[t]
\centering
\caption{Top: detectors on the same strict identity-disjoint protocol, matched sampling, 95 real and 1{,}301 fake test videos shared by all four detectors, mean of 5 seeds. Bottom: CLIP vs.\ fusion on the matched feature cache, with 95\% CIs; strict uses a fixed 30 epochs, as in Table~\ref{tab:protocols}. GF-eval follows the benchmark protocol, in which about 90\% of test reals are also used in fine-tuning; with only the official test reals, the three rows give 0.844, 0.830, and 0.513.}
\label{tab:general}
\begin{tabular}{lcccc}
\toprule
Model & FS & FR & TF & All \\
\midrule
Face CHROM & 0.877 & 1.000 & 0.992 & 0.951 \\
CLIP (frozen) & 0.971 & 0.956 & 0.943 & 0.957 \\
Xception (frozen) & 0.673 & 0.841 & 0.806 & 0.765 \\
RhythmFormer & 0.499 & 0.634 & 0.727 & 0.614 \\
\midrule
 & \multicolumn{2}{c}{Strict} & \multicolumn{2}{c}{GF-eval} \\
CLIP only & \multicolumn{2}{c}{0.957 [0.946, 0.968]} & \multicolumn{2}{c}{0.885 [0.877, 0.891]} \\
CLIP + rPPG + DCT & \multicolumn{2}{c}{0.938 [0.923, 0.953]} & \multicolumn{2}{c}{0.880 [0.872, 0.887]} \\
rPPG + DCT & \multicolumn{2}{c}{0.680 [0.633, 0.726]} & \multicolumn{2}{c}{0.547 [0.534, 0.560]} \\
\bottomrule
\end{tabular}
\end{table}

Source sensitivity is not limited to temporal traces. A frozen CLIP backbone also labels most FF++ real videos as fake (Table~\ref{tab:source}). On a common protocol, CLIP and face CHROM are comparable overall (0.957 vs.\ 0.951, with overlapping 95\% CIs), with face CHROM stronger on TF and FR and weaker on FS (Table~\ref{tab:general}). Adding rPPG and frequency features to CLIP does not help. On the strict identity-disjoint protocol the fusion scores 0.938 against 0.957 for CLIP alone, and under the benchmark's GF-eval protocol 0.880 against 0.885; across eight protocols, fusion never exceeds CLIP alone by more than 0.004. Removing the rPPG or DCT branch from the trained fusion leaves its AUC unchanged, while removing CLIP collapses it (0.555 on GF-eval). For reference, DeepFakesON-Phys \cite{hernandez2021} reaches 0.575 on our TF test set, and 0.548 with our face crops, so its weakness is not due to its face detector.

\section{Evaluation Pitfalls on the Celeb-DF Family}
\label{sec:eval}

\begin{table}[t]
\centering
\caption{Overlap between training and test in the official splits (test side). $^*$FF++ has no identity labels; its split is defined by source videos. $^\dagger$No TF subset.}
\label{tab:overlap}
\footnotesize
\setlength{\tabcolsep}{3pt}
\begin{tabular}{lccc}
\toprule
 & Celeb-DF++ & Celeb-DF-v2 & FF++ \\
\midrule
Test fakes sharing an identity & 100\% & 100\% & --$^*$ \\
FS/FR test fakes sharing a pair & 100\% & 100\% & 0\% \\
FS/FR test fakes on a training real & 78--80\% & 69\% & 0\% \\
TF test fakes from a training real & 85\% & N/A$^\dagger$ & N/A$^\dagger$ \\
Test reals used for training fakes & 92--100\% & 90\% & 0\% \\
\bottomrule
\end{tabular}
\end{table}

\subsection{Overlap in the official splits}
The official Celeb-DF and Celeb-DF++ test lists are defined per video, not per identity (Table~\ref{tab:overlap}). Every test fake shares an identity with training data, every FS and FR test fake shares its identity pair with training fakes, and most test fakes were generated from a real video in the training set. Every Celeb-DF++ test real was used as the source of a training TF fake. These overlaps are content, not only identity: for 99.7\% of test fakes, the training video they were made from is more similar in CLIP feature space than the average of the same person's other training videos. The FF++ split, defined by pairs of source videos, has no such overlap.

Whether the overlap changes reported results depends on the model and on how the overlapping item relates to the test item. We remove overlapping training items and compare against removing the same number of random items. Training on the official Celeb-DF++ training list without FF++ pretraining, removing training fakes that share an identity pair with test fakes gives CLIP 0.759, against 0.819 for random removal (paired difference $-0.061$ [$-0.085$, $-0.036$]), suggesting that this same-label overlap inflates CLIP by about 0.06. Removing training reals that served as sources for test fakes, a cross-label overlap, raises AUC relative to random removal, slightly for CLIP (+0.012) and more for a frozen Xception head (+0.040), so this kind of overlap deflates results. The official split therefore does not bias all detectors in the same direction, which makes comparisons between detectors on it harder to interpret.

\subsection{Custom protocols}

\begin{table}[t]
\centering
\caption{Leave-one-method-out (LOMO) and identity-disjoint protocols on Celeb-DF++, matched sampling. Loose: identities and real videos shared. Clean: official test reals held out. Strict: identities held out. LOMO values are the mean AUC over the 21 held-out methods.}
\label{tab:protocols}
\begin{tabular}{lccc}
\toprule
Protocol & rPPG+DCT & CLIP & Fusion \\
\midrule
Random video split & 0.680 & 0.976 & 0.974 \\
LOMO, loose & 0.929 & 0.994 & 0.996 \\
LOMO, clean reals & 0.600 & 0.971 & 0.967 \\
LOMO, strict & 0.634 & 0.947 & 0.929 \\
Identity-disjoint, strict & 0.680 & 0.957 & 0.938 \\
\bottomrule
\end{tabular}
\end{table}

Studies that build their own protocols from these datasets can introduce larger leaks. In a leave-one-method-out protocol where real test videos also appear in training, a weak rPPG+DCT detector scores 0.929 [0.925, 0.933]; holding identities out reduces this to 0.634 [0.585, 0.683], a difference of 0.30 (Table~\ref{tab:protocols}). The same change costs CLIP 0.05 (0.994 to 0.947 [0.934, 0.959]). Weak features are inflated far more than strong ones, plausibly because memorized real videos are a larger share of what they can use. Holding out only the official test reals while sharing identities gives 0.600 [0.562, 0.641] for rPPG+DCT, so the inflation comes mainly from reusing test videos. The LOMO and identity-disjoint rows use a fixed 30 epochs; with early stopping the identity-disjoint values change by less than 0.01.

\section{Discussion}

\textbf{Why animation may add flicker.} A real video records one continuous scene, so its color changes slowly. A generator that animates a still image renders each frame from the same source appearance plus a predicted motion, and nothing ties one frame's small color errors to the next. Frame-wide flicker, down to alternation between consecutive frames, is a natural result, and it fits the effect appearing when a sequence is synthesized from one frame but not when each frame is rebuilt from itself. We offer this as a hypothesis; testing it requires access to generator internals.

\textbf{Implications for deployment.} A detector that encodes video source will raise false alarms on genuine footage from unseen cameras, platforms, or processing chains, as the FF++ results show.

\textbf{Recommended controls.} Detectors motivated by physiology should report, alongside accuracy: (1) the same model on background regions and on full frames without face detection; (2) a frame-shuffle control; (3) separate performance in the heart-rate band and outside it; (4) results with real and fake videos sampled identically and re-encoded identically; (5) real-versus-real and cross-source scoring; and (6) identity-disjoint splits that account for every identity named in a filename.

\textbf{Limitations.} Without ground-truth pulse, we can show that the detectors do not rely primarily on a pulse, not that every fake lacks one. The test set has about 100 real videos from 9 identities, and the self-reenactment experiment uses two generators, 32 to 34 test clips per class, and a frame-by-frame control for one generator only. The real videos' acquisition and processing history also contributes to the cue, and the near-Nyquist excess in TF fakes is much larger than in our self-reenactments, so part of it may come from generator output or encoding pipelines rather than from animation itself. Our trace classifiers share the design of published rPPG detectors but are not those systems; DeepFakesON-Phys, the one we could run, reaches 0.575 on TF. We do not compare against audio-visual or lip-sync detectors, since our question is what rPPG-style detectors learn. Trace detectors also transfer poorly across datasets (face CHROM 0.504 from TF to FF++).

\section{Conclusion}

On the forgeries where the physiological premise is strongest, rPPG-style detectors mostly do not detect a pulse. On talking-face forgeries, a plain chrominance trace reaches 0.998 AUC from the face, 0.998 from a background region, and 0.999 from the full frame, above a dedicated rPPG extractor (0.869). Its information lies above the heart-rate band (0.992 for 4--12\,Hz alone vs.\ 0.943 for 0.7--4\,Hz), the models rely on alternation between consecutive frames, and the signal disappears when frames are shuffled (0.521). Low-passing fake traces at 8\,Hz makes 97\% of them appear real, adding fake-level 12--12.5\,Hz content to real traces makes 99.8\% appear fake, and injecting a synthetic pulse leaves only 2.5\% of fakes called real. Animating a real video from its own first frame makes it separable from its source (up to 0.931), while rebuilding each frame from itself does not (0.582), which points to frame-to-frame color flicker introduced by image-to-video synthesis.

The same detectors also encode where a video comes from. A face chrominance detector trained on Celeb-DF++ labels 99.1\% of real FF++ videos as fake; a frozen CLIP detector shows the same behavior and is not improved by adding rPPG and frequency features (0.957 alone vs.\ 0.938 fused). Evaluation practice can hide these problems: the official Celeb-DF splits share identities and source videos between training and test, and a leave-one-method-out protocol that reuses test reals in training inflates an rPPG+DCT detector from 0.634 to 0.929 AUC. Physiological detectors should be judged by controls that test their premise, not by accuracy alone.

\section*{Ethical Impact Statement}
This work uses only public research datasets (Celeb-DF, Celeb-DF++, FaceForensics++) under their terms of use and collects no new data from people; the self-reenactments were generated from existing dataset videos for analysis only. Showing what detectors rely on could help evasion, but the controls we recommend make detectors without a valid premise easier to catch.

\section*{Acknowledgment}
This work was supported by a Google Cloud research credits grant.

\end{document}